ORIGINAL RESEARCH

# Legal text classification in Korean sexual offense cases: from traditional machine learning to large language models with XAI insights

Jeongmin Lee[1,2]

Accepted: 24 March 2025

**Abstract**

The advancement of natural language processing (NLP) has expanded the application of AI-based text classification in the legal domain. However, accurately classifying legal documents remains a challenging task due to the complexity of legal texts and the subtle differences between legal categories. This study conducts a comprehensive evaluation of various legal text classification models, ranging from traditional machine learning techniques to state-of-the-art large language models (LLMs), based on ten legal categories of sexual offense precedents. The experimental results demonstrate that fine-tuning small-scale models (Small LMs) such as KLUE-BERT on legal data yields superior performance compared to large general-purpose models such as GPT-3.5 and GPT-4.0, as well as traditional machine learning models. In particular, KLUE-BERT achieved the highest accuracy of 99.3%, indicating that domain adaptation and fine-tuning play a more crucial role in legal document classification than model size alone. Furthermore, we employed explainable AI (XAI) techniques to analyze the model's predictions and conduct an in-depth review of misclassification cases. XAI-based analysis allowed us to identify key linguistic features influencing model decisions and revealed limitations in the model's ability to fully capture subtle textual cues. To further validate these findings, we leveraged KICS data, which closely resembles real-world legal case records, as a testbed to evaluate the model's generalization capabilities. The results indicate that the model struggles to interpret implicit contextual cues within legal texts. These findings emphasize the necessity for both high performance and interpretability in legal AI models. By utilizing XAI, this study proposes methods to enhance transparency in legal text classification, contributing to ongoing discussions on the reliability and practical application of AI in legal contexts. Our research suggests that AI-assisted tools can effectively support legal professionals in tasks such as legal document classification, legal information retrieval, and case assessment.

Extended author information available on the last page of the article

## 1 Introduction

In recent years, unprecedented advancements in natural language processing (NLP) have significantly transformed various domains, including the legal field. However, despite these advancements, legal text classification remains particularly challenging due to the complexity of legal language, the necessity for precise contextual interpretation, and the potential legal implications of misclassification. As legal systems generate vast amounts of unstructured textual data, such as case precedents, investigation reports, and legal statutes, ensuring accurate and interpretable AI-driven classification has become a critical issue in legal technology. Traditional approaches, such as manual coding of legal documents or classical machine learning methods, face scalability and accuracy limitations, particularly when applied to complex and evolving legal contexts.

Legal text classification presents unique challenges compared to general NLP tasks, as it requires distinguishing between highly nuanced legal terminologies and overlapping legal categories, where even minor textual variations can lead to different legal interpretations. Unlike conventional text classification problems, legal documents often contain specialized legal jargon, implicit contextual references, and domain-specific rules that complicate automated classification.

This study examines the evolution of text classification methods in the legal do-main, ranging from classical machine learning models to state-of-the-art large lan-guage models (LLMs). While classical machine learning techniques such as Support Vector Machine (SVM) and Random Forests laid the groundwork for automated legal text analysis, their limited ability to capture contextual nuances often results in suboptimal performance for complex cases. These models primarily rely on keyword-based pattern matching, bag-of-words models, and statistical representations, making them insufficient for understanding implicit contextual relationships in legal texts. As a result, they struggle to differentiate between overlapping legal categories and frequently misclassify cases with subtle linguistic variations.

On the other hand, LLMs like Bidirectional Encoder Representations from Transformers (BERT) and Generative Pre-trained Transformer (GPT) have demonstrated significant improvements in understanding legal texts, offering enhanced accuracy and efficiency. Due to their ability to process and generate human-like text, LLMs have excelled in general NLP tasks, achieving remarkable results in benchmarks such as the Turing Test and various reasoning-based evaluations. However, general-purpose LLMs, such as GPT-3.5 and GPT-4, are trained on a broad range of textual data and may not be specifically optimized for legal language understanding. This highlights the need for further evaluation of LLMs' applicability in high-stakes legal contexts, where accurately capturing legal nuances is crucial.

Amid the rapid progress in AI, its role as a tool to efficiently support legal professionals is gaining attention, particularly in high-stakes applications such as legal text classification. However, misclassification in legal cases—especially in complex domains like sexual offenses—can lead to serious legal ramifications, underscoring the need for AI models that not only achieve high accuracy but also enhance interpretability. To address this, this study systematically evaluates various machine learning and LLM-based models, comparing their effectiveness in distinguishing nuanced legal categories and minimizing classification ambiguity.

To address these challenges, this study not only evaluates classification accuracy across different models but also investigates the underlying causes of misclassification in legal text classification. Errors in classification often stem from ambiguous legal terminology, overlapping legal categories, and implicit contextual factors that conventional AI models struggle to capture. By analyzing misclassified cases, this study aims to uncover systematic weaknesses in AI-driven legal classification and identify potential strategies for improvement. Understanding these patterns is crucial for developing models that can better differentiate nuanced legal distinctions and minimize critical errors in legal text processing.

Additionally, we leverage Explainable AI (XAI) techniques to enhance model interpretability and ensure that AI-driven classifications are transparent and aligned with legal principles. While AI-based legal classification systems can achieve high accuracy, their decision-making processes remain largely opaque, limiting their trustworthiness in real-world legal applications. Through attribution analysis and word importance visualization, we examine which textual elements significantly influence classification decisions. This approach allows legal professionals to assess whether AI models are making legally sound judgments and offers a pathway tow.

Our contributions are as follow:

**A Comprehensive Analysis of Legal Text Classification and Model Interpretability.** This study conducts a comprehensive comparison of ML and LLM approaches for legal text classification in the Korean legal domain, particularly focusing on sexual offense cases. The classification models are trained and evaluated on precedent case data to ensure robust legal text processing. Additionally, KICS data, an investigative dataset, is incorporated to assess real-world applicability. By applying XAI techniques, this study examines model interpretability and decision-making patterns, providing insights into both classification performance and its potential deployment in legal practice.

**Misclassification analysis.** Legal text classification poses unique challenges, including ambiguous legal terminology, overlapping legal categories, and implicit contextual dependencies that are difficult to model using conventional approaches.

To address these challenges, we examine the underlying reasons behind recurring misclassification patterns, identifying key factors that contribute to errors. This analysis provides valuable insights into improving model robustness and ensuring more reliable AI-assisted legal decision-making.

## 2 Related work

### 2.1 Legal text analysis and AI applications

Research in the legal domain has increasingly focused on leveraging artificial intelligence (AI) to analyze legal precedents. Early efforts primarily involved transforming unstructured legal texts into structured data to predict sentencing outcomes based on legal factors and sentencing criteria (Quinsey et al. 1995). However, these approaches required manual coding of judgment elements, which is labor-intensive and limits scalability.

Recent advancements in machine learning have enabled the use of unstructured legal text data for tasks such as legal judgment prediction, where models identify key factors and predict case outcomes based on textual content (Jiang and Yang 2023; Zhong et al. 2018; Luo et al. 1707). Additional research includes similarity-based precedent searches to assist non-experts in finding relevant cases (Kim and Park 2022; Kang et al. 2021), entity recognition for extracting key information from legal documents (Kim and Lim 2022; Mistica et al. 2020), and automatic summarization to distill essential content (Yoon et al. 2022; Sheik and Nirmala 2021; Merchant and Pande 2018). These developments demonstrate the growing potential of unstructured data analysis in addressing legal challenges, laying the groundwork for this study's focus on enhanced text classification models.

In addition to case outcome prediction and information extraction, research has explored automating legal document classification to improve case retrieval and legal analytics. Various studies have employed supervised learning approaches to classify legal texts into predefined categories (Bartolo et al. 2019), while others have explored unsupervised clustering techniques for topic modeling in legal documents (Tagarelli and Simeri 2022). These approaches underscore the growing interest in legal document classification as a means to enhance accessibility and efficiency in legal research.

## 2.2 Traditional machine learning for legal text classification

Traditional machine learning models, such as Support Vector Machine (SVM) and Random Forest, have been widely used in early legal text classification tasks (Jiang and Yang 2023; Luo et al. 1707). These models offered foundational approaches for automating legal text analysis and demonstrated efficiency in handling structured data. However, they often struggled with the intricate nature of legal language, particularly when dealing with overlapping legal categories or implicit contextual dependencies. Earlier studies attempted to refine these methods through semantic annotation and feature engineering to enhance classification accuracy (Glaser and Matthes 2020), but the scalability of these approaches remained a challenge.

Challenges such as domain-specific jargon, intricate syntactic structures, and implicit legal reasoning further limited the applicability of these methods (Zhong et al. 2018). These limitations underscore the need for more sophisticated techniques capable of addressing the unique challenges presented by legal texts, motivating the exploration of advanced language models.

### 2.3 The rise of pre-trained and large language models

The introduction of Transformer architectures marked a pivotal moment in natural language processing, enabling models to better understand contextual relationships within text. Pre-trained language models (PLMs) such as BERT (Bidirectional Encoder Representations from Transformers) and GPT (Generative Pre-trained Transformer) represented significant advancements over traditional methods, offering improved performance across various NLP tasks (Devlin et al. 2019; Radford et al. 2016).

BERT's bidirectional approach allows for context to be captured from both directions, making it particularly effective in text classification tasks (Devlin et al. 2019). GPT, on the other hand, excels in generating coherent text based on unidirectional training, providing powerful tools for analyzing and synthesizing legal content (Radford et al. 2018). More recently, large language models (LLMs) such as GPT-3.5 and GPT-4 have further expanded the capabilities of PLMs by incorporating larger datasets and more parameters, leading to remarkable improvements in understanding and generalizing legal texts (OpenAI Platform-gpt-3–5-turbo 2024).

However, general-purpose LLMs often struggle with legal-specific terminology and contextual subtleties. To address these limitations, researchers have explored fine-tuning LLMs on legal corpora to improve their domain adaptability (Peric et al. 2020). Similarly, studies on legal language modeling have shown that adapting Transformers specifically for legal texts significantly enhances classification performance (Tagarelli and Simeri 2022). Despite these advances, the comparative effectiveness of general-purpose LLMs and domain-adapted models remains an open question, necessitating further empirical evaluation.

These advancements highlight the need for systematic comparisons between traditional machine learning techniques and LLMs to determine the most effective methods for legal text classification. This study contributes to this growing field by providing a comprehensive evaluation of these approaches.

### 2.4 Explainable artificial intelligence (XAI) in legal text classification

Explainable Artificial Intelligence (XAI) is a framework designed to enhance the interpretability and transparency of machine learning and deep learning models. The necessity for XAI arises from the inherent opacity of complex models, often referred to as the "black-box" phenomenon, where the reasoning behind a model's decision is not readily accessible to human users (Gunning et al. 2019). This limitation is particularly critical in high-stakes domains such as finance, healthcare, and law, where understanding the rationale behind AI-generated decisions is essential for ensuring accountability and fairness (Conrad and Branting 2018; Padovan et al. 2023).

**XAI in legal decision-making.** In the legal field, the adoption of AI for legal text classification and predictive analytics has gained traction. However, its applicability is often hindered by a lack of interpretability. Recent research has sought to address this issue by integrating XAI methodologies into legal AI systems.

Studies have investigated how legal professionals interact with explainable AI systems, with findings suggesting that interpretability enhances trust in AI-driven legal reasoning (Górski and Ramakrishna 2021). Prior research has proposed frameworks for categorizing different types of legal explanations, aligning them with policy goals to improve the comprehensibility of AI-driven decision-making (Kesari et al. 2024). Furthermore, studies have examined how legal professionals verify AI-generated classifications and assess model reliability through interpretable outputs (Nielsen et al. 2023), as well as methods to identify rationale-bearing text snippets that contribute to AI-based classifications, highlighting the importance of text-based justifications in legal AI (Mahoney et al. 2022).

Various explainability techniques have been explored to improve the transparency of LLM-based legal models. Visualization-based approaches, including Grad-CAM and attention-based attribution methods, have been used to highlight key textual elements that influence AI-generated classifications (Tagarelli and Simeri 2022; Nowosielski 2018). Additionally, semantic annotation techniques, such as zero-shot learning for legal texts, have been proposed to provide structured legal insights without requiring extensive domain-specific training (Savelka 2023). These methods contribute to enhancing model interpretability, allowing legal professionals to assess the reliability and decision-making processes of AI-driven classification systems.

While prior studies have explored explainable AI techniques in various legal contexts, there has been a limited focus on Korean legal text classification, particularly in the domain of sexual offense cases. Our study bridges this gap by integrating XAI techniques into Korean legal case classification, providing insights into both model interpretability and real-world applicability.

**XAI methodology in this study.** To enhance the interpretability of the legal text classification model, this study employs transformers-interpret (Pierse xxxx), an XAI framework specifically designed for transformer-based models. Transformers-interpret utilizes Layer Integrated Gradients (LIG) to generate attribution scores, which quantify the contribution of individual words or phrases to model predictions. Higher attribution values indicate greater influence in determining the classification outcome.

By applying transformers-interpret, this study examines the textual segments that have the greatest impact on classification results. Attribution heatmaps generated through this method highlight key linguistic features that drive the model's decision-making process. Analyzing these heatmaps helps determine whether model predictions align with legal principles or rely on unintended artifacts present in the dataset.

Beyond improving interpretability, this method enables an evaluation of potential biases in classification results. Identifying linguistic patterns that strongly influence predictions allows for an assessment of whether the model relies on spurious correlations or domain-specific artifacts. This analysis contributes to ensuring fairness and robustness in AI-assisted legal text classification, improving transparency in the decision-making process.

**Table 1** Crime case precedent datasets

| Dataset | Number of cases |
|---|---|
| Casename_classification | 1695 |
| Statute_classification | 1728 |
| ljp_criminal | 1200 |
| Total | 4623 |
| Deduplicated total | 4051 |

## 3 Data construction

The research leverages two main datasets: the Crime Case Precedent Data and the KICS Data.

### 3.1 The crime case precedent data

**Data Collection.** The Crime Cwase Precedent Data is obtained through two methods. Firstly, data from LBox, the first Korean legal language model trained on the legal corpus (Hwang et al. 2022), is utilized. This data is publicly available on Hugging Face and is sourced from LBox's case classification, statute classification, and Legal Judgment Prediction (LJP) criminal datasets.[1] The collected data statistics are summarized below (Table 1):

After filtering the dataset to focus on the top 10 most frequent offenses, a total of 3100 cases remained. Among these, indecent act by compulsion accounted for the largest portion, comprising 1811 cases, creating an imbalance in the dataset. To improve class balance and ensure sufficient representation across different crime categories, we collected additional precedent data from LegalSearch,[2] a platform for searching case precedents. The additional data primarily supplemented underrepresented categories to improve the overall balance in classification tasks. The detailed distribution of crime case data by source is provided in Appendix A.

Examples of the collected data are as follows (Table 2):

**Data preprocessing.** The crime facts of precedents often contain a mixture of information on concurrent crimes, which can complicate the classification process and hinder clear decision-making by the model. To ensure the model accurately reflects the legal context and captures the nuances of each case, we adopted single-label classification and systematically disentangled these facts. This refinement enhances the model's interpretability by allowing for a clearer understanding of its decision-making process. Therefore, the following rules were applied:

Firstly, in cases where multiple charges are involved, the crime facts are usually written with sequences (e.g., 1. 2. 3., A. B. C.) We processed them according to the method shown in Table 3. If the crime facts were not separated by charge, causing potential confusion for model training, they were excluded. Only precedents with crime facts recorded separately by charge were selected as the final data.

[1] https://huggingface.co/datasets/lbox/lbox_open

[2] https://legalsearch.kr/

**Table 2** Example of collected data

| Case 1: Indecent Act by Compulsion | Case 2: Crime of Taking Photographs or Videos by Using Cameras |
|---|---|
| **Source:** LBox | **Source:** Legal-Search |
| **Facts:** The defendant had no prior acquaintance with the victim, B (female, 21 years old)<br>On November 22, 2019, at approximately 20:35, in front of location D in Sasang-gu, Busan, the defendant forcibly molested the victim, who was working as a promotional part-timer, by grabbing her from behind with both arms | **Facts:** On March 21, 2014, at approximately 13:00, at the X Station platform on Subway Line 4 in Y-dong, Z-gu, Seoul, the defendant secretly filmed the legs of an unidentified female victim sitting on a bench using a Nokia mobile phone with a built-in camera<br>By doing so, the defendant captured the victim's body, which could provoke sexual<br>desire or shame, against her will using a camera-equipped mobile phone |

Secondly, the criminal history of the defendant, as listed in the criminal facts, was refined as it was deemed capable of biasing the data.

Third, revised clauses, such as intrusion into multiple-use places for sexual purposes (formerly intrusion into public places for sexual purposes), were integrated into more recent terminology.

After the above process, the refined precedents were sorted in descending order based on the number of cases for each offense. The top 10 offenses with sufficient data for minimum training, validation, and evaluation were determined as the final classification classes.

**Data statistics.** To divide the precedent data into train, validation, and test datasets, the data is randomly sampled in a ratio of 6:2:2. During model training, the training data is used for learning, and the validation data is utilized to monitor the progress of training. Finally, the test data is employed to conduct both quantitative and qualitative assessments of the model.

Given the potential imbalance in class proportions, stratification methods were employed to ensure that each class is evenly distributed across the train, valid, and test datasets (Kohavi 1995). The statistics for the final dataset can be found in Table 4**.**

### 3.2 KICS data

**Data description.** The Korea investigation case system (KICS) generates data during the process of case registration by investigative officers. It includes structured and unstructured data, such as personal information (e.g., names, identification numbers, contact details) and detailed crime descriptions recorded following the six interrogative principles. While KICS data holds significant potential for research, its use is restricted by the Criminal Justice Procedure Digitalization Promotion Act, which prohibits usage beyond criminal justice operations. To overcome this limitation, our study utilized de-identified, synthetic crime descriptions resembling KICS data. These crime descriptions were generated by anonymizing and pseudonymizing original legal case records, ensuring that personally identifiable information such as names, dates, and specific locations was removed or replaced. This transformation preserved the linguistic and structural characteristics of the original cases while maintaining consistency with KICS data.

**Table 3** Example of handling concurrent crimes

| Type of charges | Processing rule |
|---|---|
| Indecent act by compulsion, public indecency | Both are sexual offenses, so the crime facts are classified and included in their respective categories |
| Indecent act by compulsion, larceny | While indecent act by compulsion is considered a sexual offense, larceny is not. Therefore, the crime facts are separated, and only indecent act by compulsion is included |

**Table 4** Statistics of crime facts in sexual crime precedents dataset

| No | Crime | Train | Valid | Test | Total |
|---|---|---|---|---|---|
| 0 | Rape | 33 | 11 | 11 | 55 |
| 1 | Indecent Act by Compulsion | 1114 | 372 | 371 | 1857 |
| 2 | Public Indecency | 117 | 39 | 39 | 195 |
| 3 | Indecent Act in Crowded Public Places | 34 | 11 | 11 | 56 |
| 4 | Prostitution Arrangement | 145 | 48 | 49 | 242 |
| 5 | Quasi-Rape | 111 | 37 | 37 | 185 |
| 6 | Quasi-Indecent Act by Compulsion | 136 | 45 | 45 | 226 |
| 7 | Crime of Taking Photographs or Videos by Using Cameras | 298 | 100 | 99 | 497 |
| 8 | Obscene Act by Using Means of Communication | 119 | 40 | 40 | 199 |
| 9 | Indecent Act by Compulsion on Minors Under Age of 13 | 40 | 13 | 14 | 67 |
| | Total | 2147 | 716 | 716 | 3579 |

**Data statistics.** We extracted 246 sexual offense cases from 15,000 entries, each comprising 150–500 characters. The statistics of the dataset are as below (Table 5):

**Table 5** Statistics of KICS dataset

| No | Crime | Count |
|---|---|---|
| 0 | Rape | 32 |
| 1 | Indecent Act by Compulsion | 86 |
| 2 | Public Indecency | 10 |
| 3 | Indecent Act in Crowded Public Places | 2 |
| 4 | Prostitution Arrangement | 12 |
| 5 | Quasi-Rape | 41 |
| 6 | Quasi-Indecent Act by Compulsion | 13 |
| 7 | Crime of Taking Photographs or Videos by Using Cameras | 16 |
| 8 | Obscene Act by Using Means of Communication | 11 |
| 9 | Indecent Act by Compulsion on Minors Under Age of 13 | 23 |
| | Total | 246 |

## 4 Experiment and results

### 4.1 Baseline models

**Machine learning (ML) models**. To provide a comparative benchmark, we included traditional machine learning models such as Support vector machine (SVM) and Random Forest (RF). These models have been widely used for legal text classification.

**Small LMs (-1B)**. We considered the following models as candidates for small language models: KLUE-BERT (111M),[3] and KPF-BERT (110M),[4] both of which were pre-trained on Korean text corpora. Additionally, we included the GPT-based model LBox-lcube (124M),[5] which was fine-tuned on legal case texts and used as our baseline model to assess performance in legal text classification.

**Medium LMs (1B-10B)**. We included Llama-3.1-8B[6] and polyglot-ko-3.8b[7] as representatives of medium-sized language models. These models leverage a larger parameter space to improve contextual understanding in Korean legal text classification.

**Large LMs (10B-100B)**. For large-scale models, we utilized polyglot-ko-12.8b[8] and Llama-2-13b,[9] which provide more advanced natural language understanding capabilities due to their extensive training on multilingual and Korean-specific datasets.

3 https://huggingface.co/klue/bert-base.

4 https://github.com/KPFBERT/KPFbert/tree/main.

5 https://huggingface.co/lbox/lcube-base.

6 https://huggingface.co/meta-llama/Llama-3.1-8B.

7 https://huggingface.co/EleutherAI/polyglot-ko-3.8b.

8 https://huggingface.co/EleutherAI/polyglot-ko-12.8b.

9 https://huggingface.co/meta-llama/Llama-2-13b.

**Very Large LMs (100B+)**. GPT-3.5 (175B) and GPT-4.0 (1.8T), both of which have estimated parameter counts were included as representatives of very large-scale language models.

To ensure a robust comparison, we included LBox-lcube as a baseline model since it was trained on the same dataset used in this study. This allowed us to evaluate the performance of other models against a known benchmark and assess the impact of specialized pretraining on the dataset.

Further details regarding the experiment setup, training configurations, and ChatGPT classification instructions can be found in Appendix A.1–A.3.

### 4.2 Results of classification performance

All classification experiments were conducted using the Crime Case Precedent Data, which was used for training and evaluation. Based on the classification results in Table 6, models categorized by size, ranging from small to very large, showed varying levels of performance. Among the small language models (-1B parameters), KLUE-BERT and KPF-BERT demonstrated the highest accuracy, achieving 99.30 and 98.74% respectively. Notably, LBox-lcube, a GPT-based model partially pretrained on the same dataset, also achieved 98.74% accuracy. However, despite being tailored to the dataset, LBox-lcube did not surpass KLUE-BERT, which highlights the strength of KLUE-BERT's bidirectional architecture and its fine-tuning process in handling domain-specific tasks like legal text classification.

For medium and large language models (1B–100B parameters), such as Llama-3.1-8B, polyglot-ko-3.8b, polyglot-ko-12.8b, and Llama-2-13b, the performance ranged from 63.27 to 79.28% in accuracy. However, both medium and large models were outperformed by the fine-tuned small models, indicating that domain-specific optimization plays a crucial role in achieving high performance, regardless of model size. This result underscores that model size alone does not guarantee superior performance without appropriate fine-tuning (Table 6).

For very large-scale models (100B+parameters), GPT-3.5 and GPT-4.0 showed varied performance depending on the evaluation setting. Under Zero-Shot (ZS) conditions, GPT-4.0 outperformed GPT-3.5, achieving 62.83% accuracy compared to GPT-3.5's lower score. In Few-Shot (FS) settings, GPT-4.0's accuracy improved significantly to 72.61%, while GPT-3.5 also benefited from FS. This indicates that the more recent GPT-4.0, with its higher parameter count, is superior in its generalization capability. Moreover, the results demonstrate that providing even a small number of task-specific examples in FS settings enhances the model's ability to understand context, further boosting its performance.

Classical machine learning models, such as Support vector machine (SVM) and Random forest (RF), served as foundational baselines, achieving accuracy scores of 85.26 and 83.71%, respectively. While they provided an essential benchmark, their performance fell short of modern language models, further emphasizing the necessity of leveraging advanced techniques for nuanced and context-heavy tasks like legal text classification.

**Table 6** Classification results

| Model | Accuracy | Precision | Recall | F1 score |
|---|---|---|---|---|
| Support Vector Machine | 0.8526 | 0.8975 | 0.8790 | 0.8873 |
| Random Forest | 0.8371 | 0.8348 | 0.8201 | 0.8269 |
| KLUE-BERT | **0.9930** | <u>0.9847</u> | **0.9634** | **0.9730** |
| KPF-BERT | <u>0.9874</u> | 0.9817 | <u>0.9409</u> | <u>0.9590</u> |
| LBox-lcube | <u>0.9874</u> | **0.9944** | 0.9304 | 0.9581 |
| polyglot-ko-3.8b | 0.6327 | 0.6058 | 0.5931 | 0.5995 |
| Llama-3.1-8B | 0.7928 | 0.7325 | 0.7068 | 0.7194 |
| Polyglot-ko-12.8B | 0.7144 | 0.6952 | 0.7046 | 0.6999 |
| Llama-2-13b | 0.7851 | 0.7984 | 0.7945 | 0.7963 |
| GPT-3.5 (ZS) | 0.5942 | 0.4123 | 0.4329 | 0.4223 |
| GPT-3.5 (FS) | 0.6412 | 0.5830 | 0.5721 | 0.5775 |
| GPT-4.0 (ZS) | 0.6283 | 0.5545 | 0.5027 | 0.5273 |
| GPT-4.0 (FS) | 0.7261 | 0.6544 | 0.6823 | 0.6681 |

Bold indicates the highest performance, while underline represents the second-highest performance in each row

*GPT-3.5 and 4.0 are experimented using Zero-Shot (ZS) and Few-Shot (FS). ZS refers to tasks performed without any prior examples, while FS involves using a small number of task-specific examples for better understanding

Overall, these results highlight the significant advantage of fine-tuned, domain-specific models in legal text classification. While very large models like GPT-4.0 demonstrated promising improvements under Few-Shot conditions, the superior performance of smaller, fine-tuned models underscores the importance of tailoring models to specific tasks. This study provides a comprehensive comparison, demonstrating that even with advancements in LLMs, task-specific optimization remains crucial for achieving state-of-the-art results in specialized domains such as legal text analysis.

### 4.3 Misclassification analysis

To further assess model performance, we analyze the misclassification patterns of the best-performing model, KLUE-BERT. The confusion matrix of KLUE-BERT provides insights into the types of errors made by the model when classifying legal cases. The confusion matrix used in this analysis is normalized, meaning it represents the proportion of misclassified cases relative to the total number of cases in each category. The vertical axis represents the actual labels, while the horizontal axis represents the predicted labels. The diagonal elements indicate correctly classified cases, whereas the off-diagonal elements represent misclassifications (Fig. 1 and Table 7).

The observed misclassification patterns in KLUE-BERT reveal three key challenges in legal text classification: contextual misinterpretation of location descriptions, ambiguity between similar offense categories, and limitations in processing legal terminology. These challenges suggest that despite its high accuracy, KLUE-BERT encounters limitations in specific nuances inherent in legal texts.

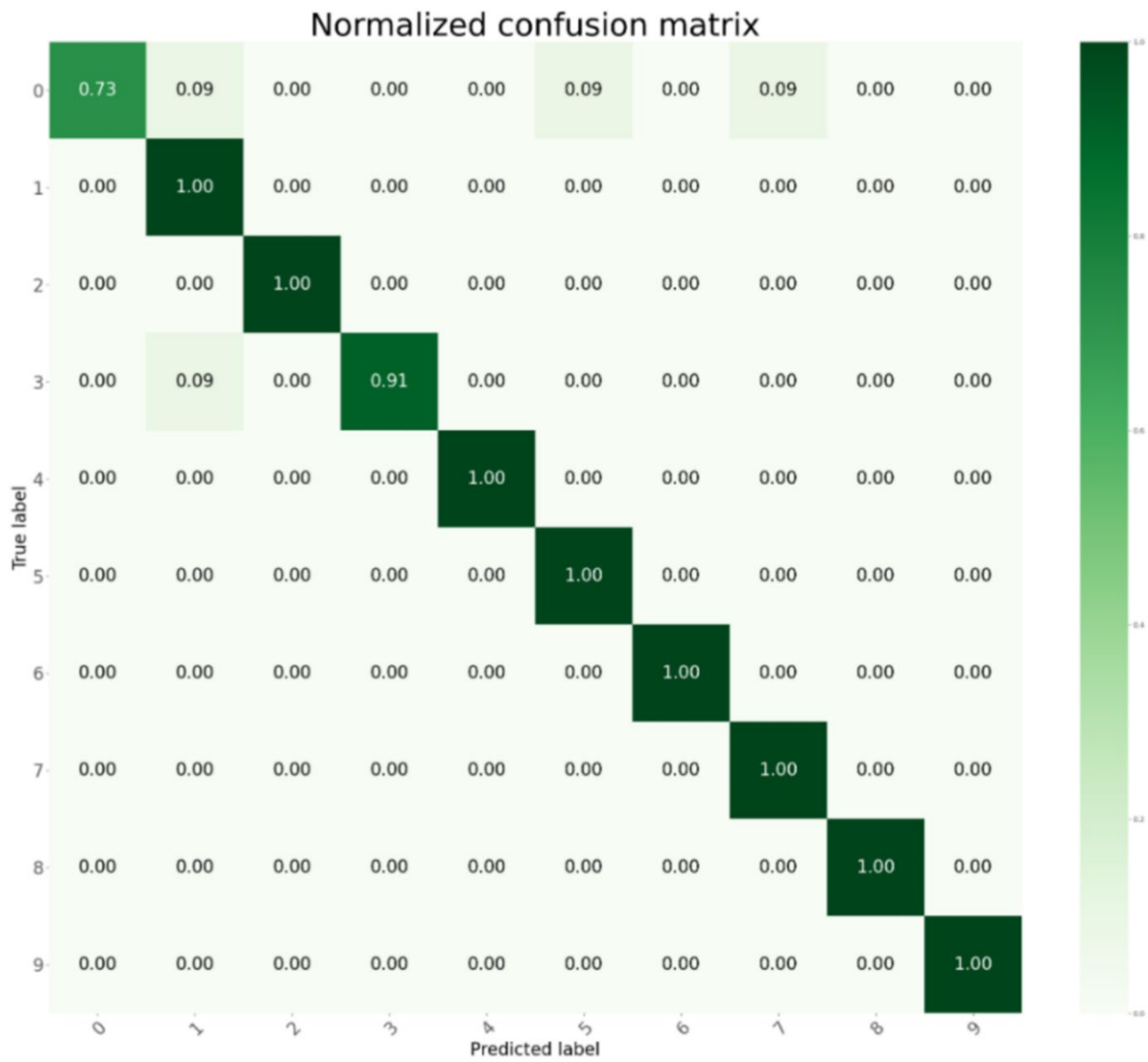

**Fig. 1** Misclassification results in KLUE-BERT

**Table 7** Misclassification cases in KLUE-BERT

| No | Actual label | Predicted label | Misclassified cases (Rate) |
|---|---|---|---|
| 1 | 0 (Rape) | 1 (Indecent Act by Compulsion) | 1 case (9%) |
| 2 | | 5 (Quasi-Rape) | 1 case (9%) |
| 3 | | 7 (Crime of Taking Photographs or Videos by Using Cameras) | 1 case (9%) |
| 4 | 3 (Indecent Act in Crowded Public Places) | 1 (Indecent Act by Compulsion) | 1 case (9%) |
| Total | | | 4 cases |

**Contextual misinterpretation of location descriptions.** A key misclassification involved Indecent Act in Crowded Public Places (label 3) being classified as

Legend: Negative Neutral Positive

| True Label | Predicted Label | Attribution Label | Attribution Score | Word Importance |
|---|---|---|---|---|
| 5 | 준강간 (1.00) | 준강간 | 4.15 | [CLS] 피의자 방 ##시 ##후 ##와 피해자 박 ##아 ##미 ##는 친구 ##관계 ##이다 .<br>피의자 ##는 피해자 ##와 성관계 를 할 목적 ##으로 2012 / 10 / 28 19 : 18 ##경 경<br>기도 김포시 감정 ##동 121 - 5 ##에 위치 ##한 " 유정 ##포 ##차 " 에서 피해자 ##<br>를 만나 같이 술 ##을 마셨 ##으며 , 같 ##은 일 20 : 30 ##경 경기도 김포시 감정 ##<br>동 137 - 25 ##에 위치 ##한 ' 장미 ##여 ##관 ' 40 ##5 ##호 ##에 피해자 ##를 끌<br>##고 입 ##실 ##하 ##였 ##다 . 피의자 ##는 술 ##에 취해 잠들 ##어 있 ##는 피해<br>자 ##를 흔들 ##어 의식 ##이 없 ##는 것 ##을 확인 ##한 후 , 피해자 ##의 속 ##바<br>지 ##와 팬티 ##를 벗기 ##고 피해자 ##의 음 ##부 ##에 성기 ##를 2 ##회 삽입 ##<br>하 ##며 피해자 ##를 간음 하 ##였 ##다 . 이 ##로 ##써 피의자 ##는 피해자 ##의 항<br>거 불능 상태 ##를 이용 ##하여 피해자 ##를 강간 하 ##였 ##다 . [SEP] |

**Fig. 2** Word Importance for Correct Classification of Quasi-Rape

Indecent Act by Compulsion (label 1). In this case, the offense occurred on a public bus (Bus No. 50), but the model failed to recognize that a bus qualifies as a crowded public place under legal definitions. This suggests that KLUE-BERT lacks robust contextual learning for location-based classifications, leading to errors when distinguishing between offenses that depend on spatial characteristics.

**Ambiguity between similar offense categories.** Several misclassified cases highlight semantic overlap between offense categories. For example, Rape (label 0) was classified as Indecent Act by Compulsion (label 1), and Quasi-Rape (label 5) was similarly misclassified. This suggests that KLUE-BERT has difficulty distinguishing offenses that share overlapping legal elements, particularly those dependent on implicit contextual factors rather than explicit wording.

To mitigate this, reinforcing Word Sense Disambiguation (WSD) techniques and contrastive learning could help improve classification by emphasizing distinguishing textual features. Incorporating context-aware embeddings trained on legal corpora may also enhance the model's ability to capture subtle distinctions between similar legal terms.

**Limitations in processing legal terminology.** While KLUE-BERT performed well overall, errors in classification indicate that the model may not fully grasp domain-specific legal language. Terms describing legal conditions, procedural contexts, or jurisdictional nuances may be processed too generally, leading to misclassification in specific cases. This highlights the need for further domain adaptation techniques tailored to legal text structures.

### 4.4 Interpreting model decisions on KICS data using XAI

To further assess the interpretability and robustness of our model, we apply XAI analysis to predictions on KICS data. While the precedent dataset consists of well-structured legal documents, the KICS dataset represents more raw, real-world legal text, which often contains variations in phrasing and less standardized legal expressions. This difference makes KICS an ideal testbed for evaluating how well the model generalizes beyond structured legal precedents.

**Correct classification: identifying key legal indicators.** By analyzing attribution scores and word importance rankings, we assess whether the model's classification decisions align with legal principles and effectively distinguish between similar offenses. Figure 2 presents a visualization of word importance, highlighting the key linguistic features that contributed to the correct classification of a given legal case within the KICS dataset. The model assigns higher attribution scores to words that are most influential in determining the crime category. The following case illustrates how these scores played a role in classifying a quasi-rape offense.

A visualization of the word importance highlights the key terms contributing to the correct classification of "quasi-rape." The most influential words included "incapacity (불능)," "resistance (항거)," "consciousness (의식)," and "intoxicated (취해)," which are crucial elements in defining quasi-rape under legal statutes. This crime category involves engaging in sexual intercourse with a person who is unable to resist due to a mental or physical condition.

Notably, words related to the victim's state, such as "fall asleep (잠들다)" and "alcohol (술)," were identified as strong indicators, suggesting that the model correctly recognized the legal distinction between quasi-rape and other sexual offenses. In contrast to cases of rape, which often emphasize "coercion" or "force," the defining factor in quasi-rape is the victim's inability to provide consent due to incapacitation. This result suggests that the model effectively leveraged contextual cues to distinguish quasi-rape from similar offenses.

**Misclassification: limitations in contextual understanding.** To further evaluate the model's decision-making process, we analyze misclassified cases to identify underlying patterns and potential sources of error. While the best-performing model demonstrated high accuracy in legal text classification, certain errors reveal systematic limitations in how the model interprets complex legal contexts. By examining these misclassified cases, we aim to uncover specific weaknesses in the model's reasoning and propose improvements to enhance its ability to distinguish between similar legal categories.

One notable misclassification occurred in a case involving Indecent Act by Compulsion on Minors Under Age of 13 (label 10), which the model incorrectly classified as Indecent Act by Compulsion (label 1). Figure 3 presents the word importance visualization for this misclassified case.

The key factor in this misclassification is that the case description did not explicitly mention the victim's age. In legal documents, human readers can infer that the

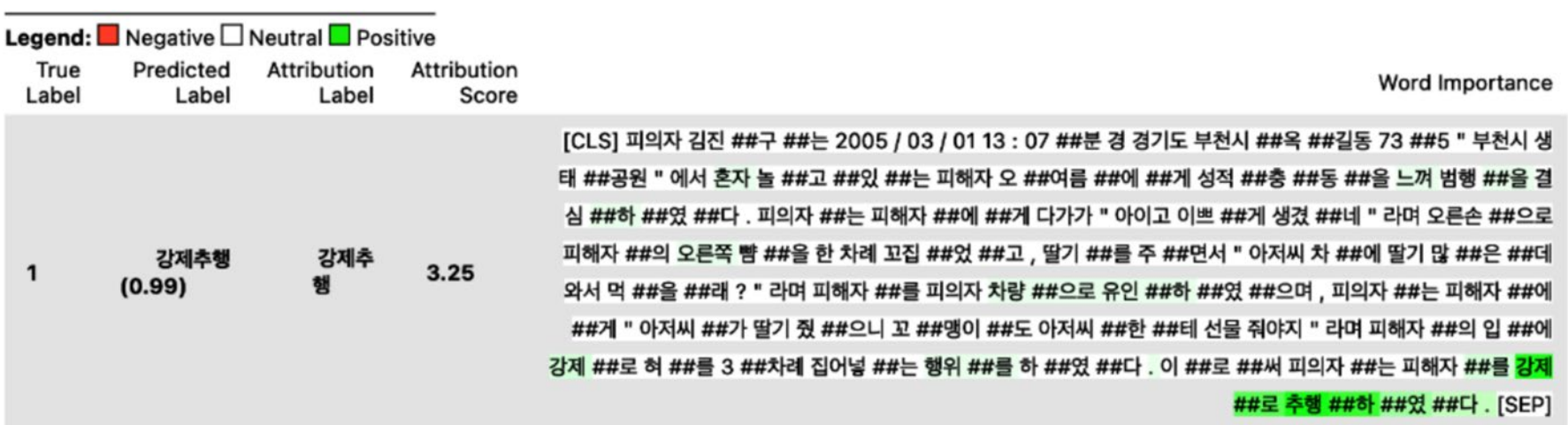

Legend: Negative Neutral Positive

| True Label | Predicted Label | Attribution Label | Attribution Score | Word Importance |
|---|---|---|---|---|
| 1 | 강제추행 (0.99) | 강제추행 | 3.25 | [CLS] 피의자 김진 ##구 ##는 2005 / 03 / 01 13 : 07 ##분 경 경기도 부천시 ##옥 ##길동 73 ##5 " 부천시 생태 ##공원 " 에서 혼자 놀 ##고 ##있 ##는 피해자 오 ##여름 ##에 ##게 성적 ##충 ##동 ##을 느껴 범행 ##을 결심 ##하 ##였 ##다 . 피의자 ##는 피해자 ##에 ##게 다가가 " 아이고 이쁘 ##게 생겼 ##네 " 라며 오른손 ##으로 피해자 ##의 오른쪽 뺨 ##을 한 차례 꼬집 ##었 ##고 , 딸기 ##를 주 ##면서 " 아저씨 차 ##에 딸기 많 ##은 ##데 와서 먹 ##을 ##래 ? " 라며 피해자 ##를 피의자 차량 ##으로 유인 ##하 ##였 ##으며 , 피의자 ##는 피해자 ##에 ##게 " 아저씨 ##가 딸기 줬 ##으니 꼬 ##맹이 ##도 아저씨 ##한 ##테 선물 줘야지 " 라며 피해자 ##의 입 ##에 강제 ##로 혀 ##를 3 ##차례 집어넣 ##는 행위 ##를 하 ##였 ##다 . 이 ##로 ##써 피의자 ##는 피해자 ##를 강제 ##로 추행 ##하 ##였 ##다 . [SEP] |

**Fig. 3** Word Importance for Misclassification of Minor Indecent Act as General Indecent Act

victim is a child based on indirect linguistic cues such as "mister (아저씨)," "luring (유인)," and "kid (꼬맹이)." These terms strongly suggest that the perpetrator targeted a minor, even without an explicit reference to the victim's age. However, the model failed to recognize these contextual signals and instead classified the case as a general indecent act without considering the victim's age.

This suggests that the model heavily relies on explicitly stated information rather than leveraging subtle contextual cues to infer victim characteristics. Unlike human readers, who can naturally interpret these social and linguistic hints, the model lacks an understanding of how certain words are commonly associated with child victims in sexual offense cases.

This limitation highlights a gap in contextual inference capability, which is crucial for accurate classification in real-world legal scenarios. A possible solution is to enhance the training dataset by incorporating more cases where indirect age-related clues are present but no explicit age reference is provided, allowing the model to learn from these patterns. Another approach could involve developing contextual embedding techniques that capture social and situational nuances beyond direct keyword matching, helping the model better recognize implicit signals. Additionally, integrating external legal knowledge bases could reinforce the model's contextual reasoning, ensuring that it correctly distinguishes between offenses that depend on inferred victim characteristics rather than explicitly stated details.

## 5 Discussion and conclusion

### 5.1 Summary of findings

This study systematically evaluated the performance of various text classification models applied to legal texts, ranging from classical machine learning approaches to large language models. The findings reveal that fine-tuned, domain-specific models outperformed both traditional machine learning classifiers, such as Support Vector Machine (SVM) and Random Forest (RF), as well as general-purpose large language models, including GPT-3.5 and GPT-4.0. This result underscores the importance of domain-specific pretraining and task-specific fine-tuning, rather than relying solely on model size or generalization capabilities. Contrary to the assumption that larger models always perform better, smaller fine-tuned models surpassed both classical machine learning models and larger LLMs. Specifically, KLUE-BERT achieved the highest classification accuracy of 99.3%, highlighting that model size alone does not guarantee superior performance in legal text classification.

The interpretability of model decisions was further analyzed using XAI techniques, uncovering both strengths and limitations. In correctly classified cases, the model effectively identified key legal terms that define specific crimes. However, misclassified cases revealed the model's difficulty in recognizing implicit contextual cues, particularly in cases where critical information about victims or circumstances was not explicitly stated. One notable example was the failure to infer that a victim was a minor based on indirect linguistic signals such as "mister," "luring," and "kid," leading to incorrect classification of Indecent Act by Compulsion on Minors Under Age of 13 as a general Indecent Act by Compulsion. Unlike human readers, who can naturally interpret social and linguistic context, the model relied heavily on explicit keywords, underscoring a key limitation in contextual reasoning.

### 5.2 Limitations

Despite the promising results, this study has several limitations. First, data imbalance remains a challenge, as certain crime categories were significantly more frequent than others, potentially biasing the model's classification tendencies. While fine-tuning improved performance, further work is needed to ensure that rare offense categories are classified with the same accuracy as more common ones.

Additionally, the study was limited to ten major categories of sexual offenses. While this selection was necessary due to dataset constraints, real-world legal applications require broader classification frameworks that account for a more extensive range of criminal offenses. Expanding the dataset and increasing category diversity would improve the generalizability of the model.

Another key limitation is the model's reliance on explicit information rather than implicit contextual reasoning. Many legal cases require interpretation of unstated or inferred elements, and current models struggle to account for such nuances. The inability to distinguish between offenses based on contextual clues, such as implicit victim characteristics, highlights a gap in AI-driven legal text classification.

### 5.3 Future works

To address these limitations, future research should explore methods to enhance contextual reasoning in legal text classification. One potential approach is to integrate reinforcement learning techniques such as Direct preference optimization (DPO) (Rafailov et al. 2024) to refine the model's decision-making process based on nuanced legal distinctions. By incorporating preference-based learning, the model could be trained to better differentiate between similar offenses, such as distinguishing between Indecent Act by Compulsion and Quasi-Indecent Act by Compulsion, or recognizing implicit age-related factors in cases involving minors.

Improving contextual embedding techniques is another promising direction. By refining language models to better capture the relationships between legal concepts and contextual cues, AI systems could move beyond keyword-based classification to a more holistic understanding of legal narratives. Additionally, integrating external legal knowledge bases could reinforce the model's ability to infer unstated but legally significant elements, ensuring more accurate case classification.

### 5.4 Implications and conclusions

The findings of this study suggest that AI-driven legal text classification models have the potential to serve as decision-support tools in legal practice. Rather than replacing human judgment, these models can enhance the efficiency of legal document classification and retrieval. Integrating such models into case management systems or legal research platforms could improve accessibility and organization of legal information.

A key contribution of this research is the application of explainable AI (XAI) techniques to legal text classification. By examining both correctly classified and misclassified cases, this study provides insights into how AI models interpret legal text and where their limitations lie. This transparency is crucial for building trust in AI-assisted legal decision-making.

Despite the high accuracy achieved by fine-tuned models, this study highlights the challenges of contextual inference in legal classification. The model exhibited misclassification errors when critical legal attributes—such as the victim's age—were inferred rather than explicitly stated. These findings underscore the need for improving models' ability to process implicit contextual cues.

To further advance AI applications in legal practice, human-AI collaboration will be essential. Expert feedback loops can be used to iteratively refine classification models, ensuring they align more closely with legal reasoning. Additionally, integrating AI models with structured legal databases and precedent analysis tools could further enhance their practical utility.

Ultimately, this study demonstrates that AI models, when combined with explainability techniques and domain expertise, can contribute to more transparent, interpretable, and effective legal text classification systems. Future research should continue refining these models to address contextual limitations, ensuring that AI-assisted legal tools remain reliable, accountable, and aligned with real-world legal practices.

## Appendix A: Distribution of crime case data by source

See Table 8.

**Table 8** Distribution of crime case data by source

| Index | Crime | Lbox | Legal-search | Total |
|---|---|---|---|---|
| 0 | Rape | 47 | 8 | 55 |
| 1 | Indecent Act by Compulsion | 1,810 | 47 | 1,857 |
| 2 | Public Indecency | 180 | 15 | 195 |
| 3 | Indecent Act in Crowded Public Places | 46 | 10 | 56 |
| 4 | Prostitution Arrangement | 213 | 29 | 242 |
| 5 | Quasi-Rape | 182 | 3 | 185 |
| 6 | Quasi-Indecent Act by Compulsion | 218 | 8 | 226 |
| 7 | Crime of Taking Photographs or Videos by Using Cameras | 439 | 58 | 497 |
| 8 | Obscene Act by Using Means of Communication | 195 | 4 | 199 |
| 9 | Indecent Act by Compulsion on Minors Under Age of 13 | 45 | 22 | 67 |
| | Total | 3375 | 204 | 3579 |

## Appendix A.1: Experiment setup

The fine-tuning process was conducted using PyTorch and Hugging Face Transformers. We employed transfer learning, initializing models with pre-trained weights and fine-tuning them on our dataset. Training was performed on two NVIDIA RTX A6000 GPUs with mixed-precision training enabled for efficiency. The optimal checkpoint was determined using early stopping based on validation loss. To ensure clarity and reproducibility, the hyperparameter settings used during fine-tuning are presented in Appendix A.2.

## Appendix A.2: Training arguments configuration

Table 9 presents the hyperparameters used in the training process.

**Table 9** Training arguments configuration

| Training arguments | Value |
|---|---|
| Initial learning_rate | 2e-5 |
| Optimizer | adamw_torch |
| Weight_decay | 0.01 |
| Loss function | Cross-Entropy Loss |
| Per_device_train_batch_size | 8 |
| Per_device_eval_batch_size | 8 |
| Num_train_epochs | 5 |
| Early stopping patience | 5 |
| Max_length | 512 |
| Save_total_limit | 3 |

## Appendix A.3: Classification instruction for ChatGPT

In this study, we conducted Zero-Shot (ZS) and Few-Shot (FS) experiments using GPT-3.5 and GPT-4 for legal text classification. The models were prompted to classify sexual offense cases into predefined legal categories based on textual descriptions of crime facts.

**Zero-Shot Setting (ZS):** No examples were provided in the prompt. The model was expected to infer the classification task solely from the given instructions.

**Few-Shot Setting (FS):** A total of 10 examples (one case per crime category) were randomly selected from the training dataset and included in the prompt as demonstrations.

The following is the exact instruction used for classification:

```
### instruction
The following is a case of a sexual offense. Please select the most appropriate category from the list provided below.

### Categories:
    1.  rape
    2.  indecent act by compulsion
    3.  public indecency
    4.  indecent act in crowded public places
    5.  prostitution arrangement
    6.  quasi-rape
    7.  quasi-indecent act by compulsion
    8.  crime of taking photographs or videos by using cameras
    9.  obscene act by using means of communication
    10. indecent act by compulsion on minors under age of 13

### Rules:
- The response must be **a number only**. For example: "1", "2", "3"
- Choose exactly one category from the list above.
- Select the category that best matches the content of the case.
- There are no cases that do not fit into the provided categories.

### Example:
facts: {facts}
casename: {casename}

### Case:
A. {}
```

**Acknowledgements** This work was supported by Institute of Information and Communications Technology Planning and Evaluation(IITP) grant funded by the Korea government(MSIT) (RS-2019-II190004, Development of semi-supervised learning language intelligence technology and Korean tutoring service for foreigners).

**Publisher's Note** Springer Nature remains neutral with regard to jurisdictional claims in published maps and institutional affiliations.

## Authors and Affiliations

### Jeongmin Lee[1,2]

✉ Jeongmin Lee
faraway@ust.ac.kr; faraway@etri.re.kr

1 University of Science and Technology (UST), Daejeon, Republic of Korea

2 Electronics and Telecommunications Research Institute (ETRI), Daejeon, Republic of Korea